\documentclass[10pt,twocolumn,letterpaper]{article}

\usepackage[pagenumbers]{cvpr}

\usepackage{graphicx}
\usepackage{amsmath}
\usepackage{amssymb}
\usepackage{booktabs}
\usepackage{multirow}
\usepackage{arydshln}

\usepackage[pagebackref,breaklinks,colorlinks]{hyperref}
\usepackage[capitalize]{cleveref}
\crefname{section}{Sec.}{Secs.}
\Crefname{section}{Section}{Sections}
\Crefname{table}{Table}{Tables}
\crefname{table}{Tab.}{Tabs.}

\def\cvprPaperID{1980} 
\def\confName{CVPR}
\def\confYear{2023}

\begin{document}

\title{Better ``CMOS" Produces Clearer Images: \\
Learning Space-Variant Blur Estimation for Blind Image Super-Resolution}
\author{Xuhai Chen$^1$\thanks{Equal contribution.}
~ ~ Jiangning Zhang$^{2}$\footnotemark[1]
~ ~ Chao Xu$^1$
~ ~ Yabiao Wang$^2$
~ ~ Chengjie Wang$^2$
~ ~ Yong Liu$^1$\thanks{Corresponding author.} \\
\normalsize $^1$ APRIL Lab, Zhejiang University ~ ~ $^2$Youtu Lab, Tencent \\
{\tt\small \{22232044, 21832066\}@zju.edu.cn, yongliu@iipc.zju.edu.cn} \\
{\tt\small \{vtzhang, caseywang, jasoncjwang\}@tencent.com}
}
\maketitle

\begin{abstract}
    Most of the existing blind image Super-Resolution (SR) methods assume that the blur kernels are space-invariant. However, the blur involved in real applications are usually space-variant due to object motion, out-of-focus, \etc, resulting in severe performance drop of the advanced SR methods. To address this problem, we firstly introduce two new datasets with out-of-focus blur, \ie, NYUv2-BSR and Cityscapes-BSR, to support further researches of blind SR with space-variant blur. Based on the datasets, we design a novel \textbf{C}ross-\textbf{MO}dal fu\textbf{S}ion network (CMOS) that estimate both blur and semantics simultaneously, which leads to improved SR results. It involves a feature Grouping Interactive Attention (GIA) module to make the two modalities interact more effectively and avoid inconsistency. GIA can also be used for the interaction of other features because of the universality of its structure. Qualitative and quantitative experiments compared with state-of-the-art methods on above datasets and real-world images demonstrate the superiority of our method, \eg, obtaining PSNR/SSIM by +1.91$\uparrow$/+0.0048$\uparrow$ on NYUv2-BSR than MANet\footnote{\url{https://github.com/ByChelsea/CMOS.git}}.
\end{abstract}

\vspace{-1.3em}
\section{Introduction}
\label{sec:intro}

Blind image SR, with the aim of reconstructing High-Resolution (HR) images from Low-Resolution (LR) images with unknown degradations, has attracted great attention due to its significance for practical use~\cite{gu2019blind, liang2021swinir, ji2020real, niu2020single, chen2022activating, liang2021mutual, kim2021koalanet, liang2021flow, fang2022uncertainty}. Two degradation models, bicubic downsampling~\cite{timofte2017ntire} and traditional degradation~\cite{liu2013bayesian, shocher2018zero}, are usually used to generate LR images from HR images. The latter can be modeled by:
\begin{equation}
  \begin{aligned}
    \boldsymbol{y}=(\boldsymbol{x} \bigotimes \boldsymbol{k})\downarrow _s+\boldsymbol{n}.
  \end{aligned}
  \label{eq:1}
\end{equation}

\begin{figure}[htp]
    \centering
    \includegraphics[width=8.2cm]{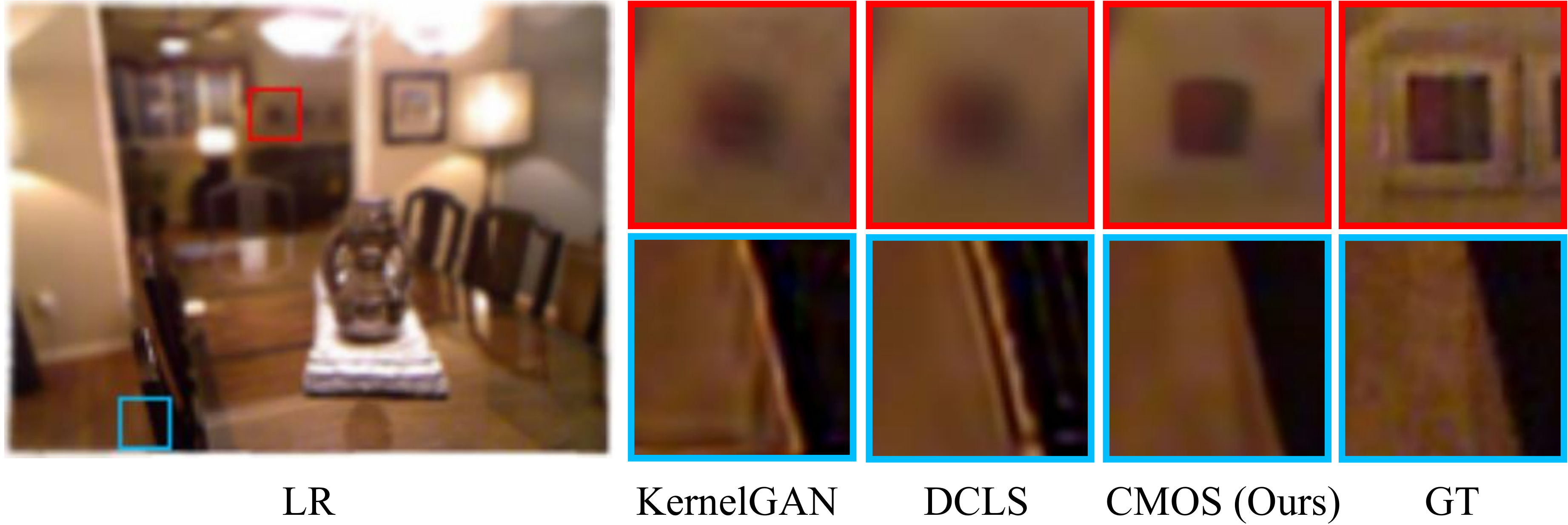}
    \caption{SR results of KernelGAN~\cite{bell2019blind}, DCLS~\cite{luo2022deep} and the proposed CMOS on a space-variant blurred LR image. For KernelGAN and DCLS, patches are blurry in the first row and have artifacts in the second row, while CMOS performs well in both cases.}
    \label{fig:introduction1}
\end{figure}

\noindent
It assumes the LR image $\boldsymbol{y}$ is obtained by first convolving the HR image $\boldsymbol{x}$ with a blur kernel $\boldsymbol{k}$, followed by a downsampling operation with scale factor $s$ and an addition of noise $\boldsymbol{n}$. On top of that, some works~\cite{realesrgan, zhang2021designing} propose more complex and realistic degradation models, which also assume that blur is space-invariant. However, in real-world applications, blur usually changes spatially due to factors such as out-of-focus and object motion, so that the mismatches will greatly degrade the performance of existing SR methods. Fig.~\ref{fig:introduction1} gives an example when the LR image suffers from space-variant blur. Since both KernelGAN~\cite{bell2019blind} and DCLS~\cite{luo2022deep} estimate only one blur kernel for an image, there are a lot of mismatches. In the first row of Fig.~\ref{fig:introduction1}, where the kernel estimated by the two methods are sharper than the real one of the patch, SR results are over smoothing and high frequency textures are significantly blurred. In the second row, where the kernels estimated are smoother than the correct one, SR results show ringing artifacts caused by over-enhancing high-frequency edges. This phenomenon illustrates that mismatch of blur will significantly affect SR results, leading to unnatural outputs. In this paper, we focus on the space-variant blur estimation to ensure that the estimated kernel is correct for each pixel in the images.

\begin{figure}[t!]
    \centering
    \includegraphics[width=8.2cm]{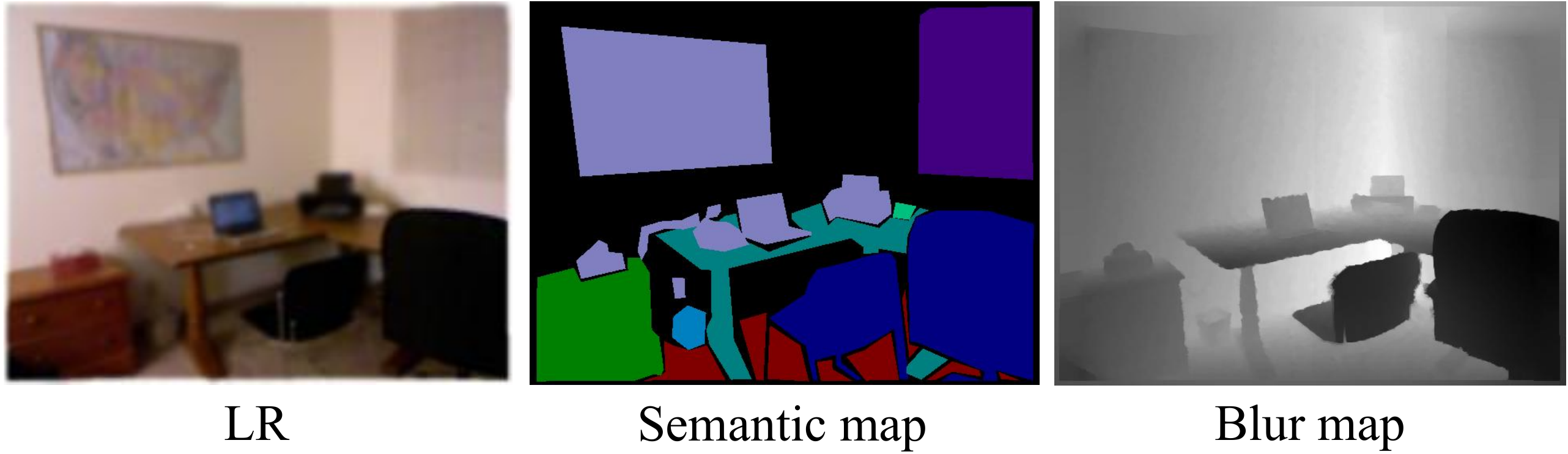}
    \caption{A condition in which blur and semantic information are inconsistent. This image comes from our dataset NYUv2-BSR.}
    \label{fig:introduction2}
\end{figure}

A few recent works~\cite{liang2021mutual, kim2021koalanet, xu2021edpn} have taken space-variant blur into account. Among them, MANet~\cite{liang2021mutual} is the most representative model, which assumes that blur is space-invariant within a small patch. Based on this, MANet uses a moderate receptive field to keep the locality of degradations. However, there are still two critical issues. 1) Because there is no available dataset containing space-variant blur in SR field, MANet is trained on space-invariant images, resulting in blur deviation of the training and testing phase. 2) Even limiting the size of the receptive field, the estimation results are still poor at the boundaries of different kernels, leading to mean value prediction of space-variant blur.

To address the aforementioned challenges, we first introduce a new degradation method and propose two corresponding datasets, \ie, NYUv2-BSR and Cityscapes-BSR to support relevant researches of space-variant blur in the SR domain. As a preliminary exploration, out-of-focus blur is studied as an example in this paper and it is generated according to the depth of the objects using the method proposed in~\cite{lee2019deep}. Besides, we also add some space-invariant blur into the datasets so that the models trained on them can cope with both spatially variant and invariant situations. 

Furthermore, to improve the performance at the boundaries of different blur regions, we present a novel model named \textit{Cross-MOdal fuSion network} (CMOS). Our intuition is that the sharp semantic edges are usually aligned with out-of-focus blur boundaries and it can help to distinguish different blur amounts. This raises a critical concern that how to effectively introduce semantics into the process. Specifically, we firstly predict blur and semantics simultaneously instead of using the semantics as an extra input, which not only avoids using extra information during test phase, but also enables non-blind SR methods to recover finer textures with the two modalities. Secondly, to enhance accuracy at the blur boundaries, we conduct interaction between the semantic and blur features for complementary information learning inspired by multi-task learning~\cite{vandenhende2020mti, xu2018pad}. However, in some cases these two modalities are inconsistent. As shown in Fig.~\ref{fig:introduction2}, the wall and the picture on it are completely different in the semantic map, with clear boundaries. But the depth of them are almost the same, so the blur amounts depending on depth are also very similar. In this case, not only can the two modalities fail to use common features, but they can also negatively influence each other. Besides, since we add some space-invariant blurred images with uniform blur maps in the datasets, it will also greatly increase the inconsistency. 

Motivated by these observations, we propose a feature Grouping Interactive Attention (GIA) module to help the interaction of the two modalities. GIA has two parallel streams: one operating along the spatial dimension and the other along the channel dimension. Both streams employ group interactions to process the input features and make adjustments. Moreover, GIA has an upsampling layer based on the flow field~\cite{li2020semantic} to support inputs of different resolutions. Its universal structure allows it to be used for more than just interactions between the two modalities.

The main contributions of this work are as follows:
\begin{itemize}
	\item To support researches on space-variant blur in the field of SR, we introduce a new degradation model of out-of-focus blur and propose two new datasets, \ie, NYUv2-BSR and Cityscapes-BSR.
	\item We design a novel model called CMOS for estimating space-variant blur, which leverages extra semantic information to improve the accuracy of blur prediction. The proposed GIA module is used to make the two modalities interact effectively. Note that GIA is universal and can be used between any two features.
	\item Combined with existing non-blind SR models, CMOS can estimate both space-variant and space-invariant blur and achieve SOTA SR performance in both cases.
\end{itemize}

\section{Related Work}
\subsection{Degradation Model}  

SR methods give rise to poor performance if the assumed degradation deviates from those in reality. Many works~\cite{zhang2018learning, elad1997restoration, yuan2018unsupervised} use the traditional model (Eq.~\ref{eq:1}) to generate their training data. Compared to bicubic downsampling~\cite{wang2018esrgan, zhao2020efficient}, although traditional model has taken more factors into account, it is still too simple to simulate real degradation. Consequently, Real-ESRGAN~\cite{realesrgan} proposes a flexible high-order degradation model by applying traditional model repeatedly, while BSRGAN~\cite{zhang2021designing} adjusts the degradation order of the traditional model and use randomly shuffled blur, downsampling and noise. Liang \textit{et al.}~\cite{liang2021mutual} go a step further to simulate space-variant blur by dividing images into patches and applying different kernels. Unfortunately, it cannot well simulate the real situations. As a result, to support relevant researches, we introduce space-variant out-of-focus blur into SR, and propose two corresponding datasets, \ie, NYUv2-BSR and Cityscapes-BSR.

\begin{figure*}[t]
    \centering
    \includegraphics[width=17cm]{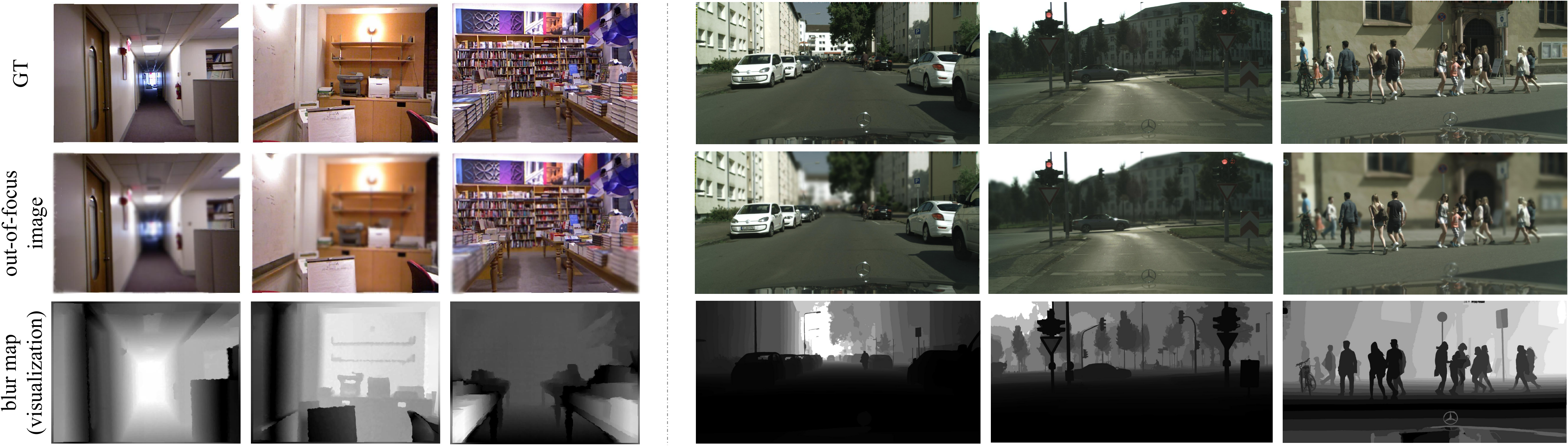}
    \caption{Original RGB images, the generated out-of-focus images and blur maps. The changes from dark to light in blur maps indicate that the corresponding out-of-focus image changes from clear to blur. The first three columns are images from NYUv2-BSR, and the last three columns are images from Cityscapes-BSR.}
    \label{fig:image example}
\end{figure*}
\vspace{0pt}

\subsection{Kernel Estimation}  
One of the mainstream methods of blind SR is to estimate degradation first and then use it as prior information for non-blind SR. KernelGAN~\cite{bell2019blind} proposes to learn a kernel from the internal distribution of image patches, while IKC~\cite{gu2019blind} uses an iterative correction scheme to learn the PCA features of kernels. Luo \textit{et al.}~\cite{luo2022deep} transfer blur estimation into LR space and learn kernel weights instead of kernel itself. However, these methods only estimate a unique kernel, thus the performance will be significantly reduced on space-variant situations. Accordingly, KOALAnet~\cite{kim2021koalanet} proposes to learn specific kernels for each pixel, and MANet~\cite{liang2021mutual} designs a network with moderate receptive field to adapt to the locality of degradation. However, they still have limitations, such as the moderate receptive field might limit the capacity of the model. By contrast, with the help of semantic information, our CMOS can predict space-variant blur effectively and accurately.

\subsection{Non-blind SR}
Non-blind SR aims to restore images with known degradations. Early non-blind SR methods~\cite{kim2016accurate, lim2017enhanced, johnson2016perceptual, ledig2017photo} are based on bicubic downsampling, which struggle to generalize to images with more complex degradations. To address this problem, SRMD~\cite{zhang2018learning} first proposes to stretch the blur and noise to the size of LR images, and take the concatenated images and degradation maps as input to restore the HR counterparts. Following SRMD, SFTMD~\cite{gu2019blind} uses SFT layer~\cite{wang2018recovering} to combine the stretching degradation maps instead of simply concatenation, while UDVD~\cite{xu2020unified} employs per-pixel dynamic convolution to more effectively deal with variational degradations across images. Besides, zero-shot methods~\cite{shocher2018zero, hussein2020correction, soh2020meta} have also been investigated in non-blind SR with multiple degradations. What is noteworthy is that our CMOS can be easily combined with most non-blind SR methods to achieve excellent blind SR performance.

\section{The Proposed Datasets}
\label{sec:sec3}

\begin{figure*}[htp]
    \centering
    \includegraphics[width=17cm]{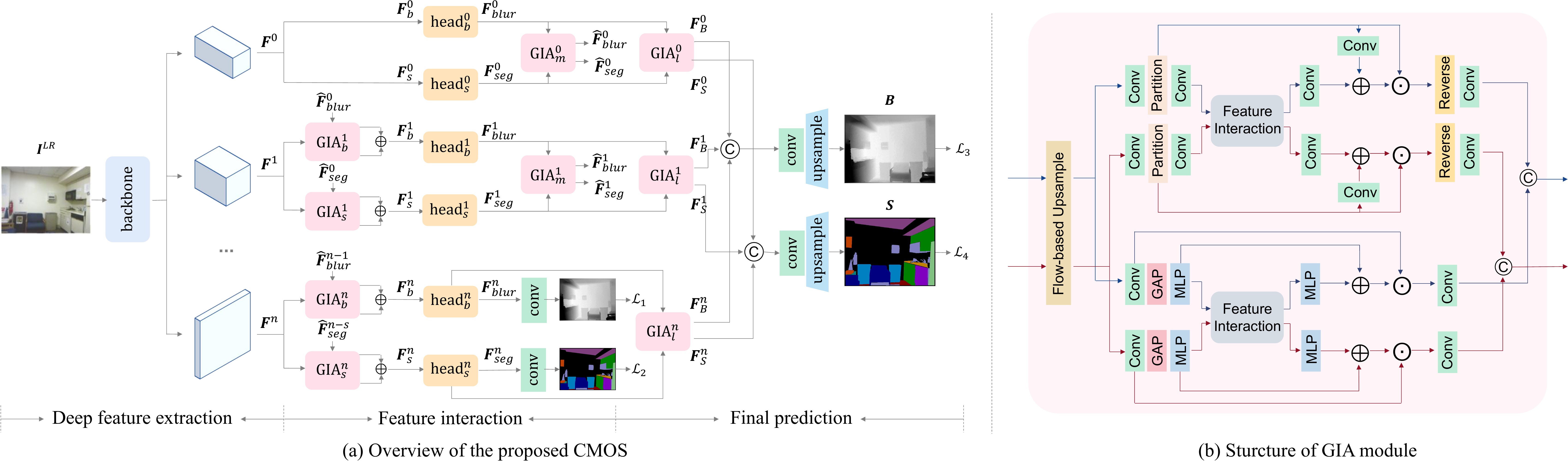}
    \caption{\textbf{Architecture of CMOS and GIA.} (a) Given an LR image, CMOS outputs the estimated blur map $\boldsymbol{B}$ and semantic map $\boldsymbol{S}$ simultaneously in the HR space. (b) GIA has two parallel streams to effectively interact features in both spatial and channel dimensions. It includes a flow-based upsample module to support inputs with different resolutions. If the input resolutions are the same, a feature alignment will also be performed through the learned flow field.}
    \label{fig:main}
\end{figure*}
\vspace{0pt}

To support researches on space-variant blur, we propose two novel datasets, NYUv2-BSR and Cityscapes-BSR, where BSR stands for Blind image SR. To the best of our knowledge, we are the first to introduce out-of-focus, one of the most common space-variant blur in real world, into blind image SR. Out-of-focus is caused by differences in depth. Every point that is not in the plane of focus corresponds to a Circle Of Confusion (COC) in image plane. The blur can be simulated by isotropic Gaussian kernels with standard deviation $\sigma$ related to the diameters of COCs~\cite{kraus2007depth}, which can be calculated using thin lens model~\cite{potmesil1981lens}. We employ the method proposed in ~\cite{lee2019deep} to blur the images and the ground truth blur map is constructed by $\sigma$ of each pixel.

\begin{table}
	\centering
    \renewcommand{\arraystretch}{1.2}
	\begin{tabular}{|c|c|c|c|c|c|c|}
		\hline
		\multirow{2}{*}{Dataset} & \multicolumn{3}{|c|}{NYUv2-BSR} & \multicolumn{3}{|c|}{Cityscapes-BSR} \\
		\cline{2-7}
		 ~ & VA & IVA & Total & VA & IVA & Total \\
		\hline
		Train & 636 & 159 & 795 & 2380 & 595 & 2975 \\
		\hline
		Val & - & - & - & 400 & 100 & 500 \\
		\hline
		Test & 524 & 130 & 654 & 1220 & 305 & 1525\\
		\hline
	\end{tabular}
	\caption{Details of NYUv2-BSR and Cityscapes-BSR. VA and IVA represents the number of images with space-variant out-of-focus blur and space-invariant blur respectively.}
    \label{tab:datasets}
\end{table}

As mentioned above, we need depth-color image pairs to generate images with out-of-focus blur. Thus, we select NYUv2~\cite{silberman2012indoor} and Cityscapes~\cite{cordts2016cityscapes} as original datasets. NYUv2 is an indoor dataset. It contains 1449 pairs of RGB and depth images, in which 795 pairs are used for training and the rest 654 for testing. Cityscapes is an outdoor dataset and the fine-annotated part consists of training, validation and test sets containing 2975, 500, and 1525 images, respectively. Since the depth maps in Cityscapes contain invalid measurements, which are not conducive to the generation of out-of-focus images, we use CREStereo~\cite{li2022practical}, a deep learning-based stereo matching method, to generate disparity maps and calculate the corresponding depth maps based on the camera parameters. Fig.~\ref{fig:image example} shows the original RGB images of NYUv2 and Cityscapes, as well as the generated out-of-focus images and corresponding blur maps.

In terms of parameters of the isotropic Gaussian kernels, the kernel width range is set to $\left[0.0, 5.0\right]$ and $\left[0.0, 15.0\right]$ for NYUv2 and Cityscapes, respectively. The kernel size is fixed to $21 \times 21$ and $61 \times 61$, and the downsampling scale factor is set to 4. Besides, 1/4 of the images are blurred by space-invariant kernels, so that the models trained on the datasets are not limited by the space-variant situations. Tab.~\ref{tab:datasets} shows the details. In addition, to ensure the adequacy and fairness of experiments, we created five test groups for each dataset, in which each group had a different 1/4 of the images blurred by space-invariant kernels.

\section{Method}

As stated before, sharp semantic edges can increase the accuracy of space-variant blur estimation near the boundaries. Motivated by this, we propose a Cross-MOdal fuSion network (CMOS) to predict both blur and semantic maps simultaneously by mutual supervision of them.

\subsection{Overview}
Inspired by \cite{vandenhende2020mti}, CMOS is a multi-scale network, which consists of three main stages, as shown in Fig.~\ref{fig:main}. In the first stage, a fully convolutional encoder capable of generating multi-scale features is used to extract deep features $\left\{\boldsymbol{F}^0, \boldsymbol{F}^1, \cdots, \boldsymbol{F}^n \right\}$. In the next stage, for each scale $i$, we apply two task-specific heads, $\text{head}_b^i$ and $\text{head}_s^i$, to predict initial blur and semantic features $\boldsymbol{F}_{blur}^i$ and $\boldsymbol{F}_{seg}^i$. Then, we use a proposed $\text{GIA}_m^i$ module to achieve effective information interaction between the two modalities to obtain more accurate features $\hat{\boldsymbol{F}}_{blur}^i$ and $\hat{\boldsymbol{F}}_{seg}^i$ in a mutually supervised manner, formulated as:
\begin{eqnarray}
    &\boldsymbol{F}_{blur}^i={\text{head}_b^i} (\boldsymbol{F}_b^i),\\
    &\boldsymbol{F}_{seg}^i={\text{head}_s^i} (\boldsymbol{F}_s^i),\\
    &\hat{\boldsymbol{F}}_{blur}^i, \hat{\boldsymbol{F}}_{seg}^i={\text{GIA}_m^i}(\boldsymbol{F}_{blur}^i, \boldsymbol{F}_{seg}^i),
\end{eqnarray}

\noindent
where $\boldsymbol{F}_b^i$ and $\boldsymbol{F}_s^i$ denotes the input of the task-specific heads. To make better use of the multi-scale information, we use $\text{GIA}_b^i$ and $\text{GIA}_s^i$ to fuse the adjacent low-scale features, so the input of the heads can be written as:
\begin{eqnarray}
    &\boldsymbol{F}_b^0 = \boldsymbol{F}_s^0 = \boldsymbol{F}^0,\\
    &\boldsymbol{F}_b^i = {\text{Sum}}({\text{GIA}_b^i}(\boldsymbol{F}^i, \hat{\boldsymbol{F}}_{blur}^{i-1})),\\
    &\boldsymbol{F}_s^i = {\text{Sum}}({\text{GIA}_s^i}(\boldsymbol{F}^i, \hat{\boldsymbol{F}}_{seg}^{i-1})),
\end{eqnarray}

\noindent
where $\text{Sum}( \cdot )$ represents for adding outputs of the modules. At the highest resolution $n$, the task-specific features are fed into two convolution layers to generate auxiliary blur and semantic maps for additional supervision, which is beneficial to further improve the accuracy of the final prediction.

The last stage consists $n+1$ $\text{GIA}_l^i$ modules to get the final features $\boldsymbol{F}_B^i$ and $\boldsymbol{F}_S^i$ of each scale as:
\begin{equation}
  \begin{aligned}
    \boldsymbol{F}_B^i, \boldsymbol{F}_S^i={\text{GIA}_l^i}(\boldsymbol{F}_{blur}^i, \boldsymbol{F}_{seg}^i).
  \end{aligned}
\end{equation}

\noindent
These features are then concated and convolved to obtain the prediction of blur and semantic maps. In this way, we can build a shorter way for each scale to the supervision and further facilitate the interaction between blur and semantics. Besides, since blur is done in the HR space, we upsample the outputs using bi-linear interpolation by scale factor $s$.


\subsection{Grouping Interactive Attention Module}
\label{sec:4.2}

GIA is designed to help blur and semantics interact more effectively and avoid inconsistency. Besides, it can also be used for other features because of the universal structure. GIA involves two parallel streams operating on spatial and channel dimensions, and it can handle inputs of different resolutions by using a flow-based upsample module~\cite{li2020semantic}.

\noindent
\textbf{Spatial Grouping Feature Interaction.}  The input features may be similar on most patches, but different on some. As shown in Fig.~\ref{fig:introduction2}, the picture hanging on the wall brings difference between the blur and semantic maps. As a result, we propose to adjust the spatial weight map in the general spatial attention~\cite{guo2022attention, zhang2021analogous, zhang2022eatformer} mechanism to take advantage of similar information and avoid inconsistencies.

In the top half of Fig.~\ref{fig:main} (b), each input is first passed through a convolution layer and divided into windows denoted by $\boldsymbol{F}_w^j$. These windows are then further processed by another convolution layer before being fed into the feature interaction module (last part of Sec.~\ref{sec:4.2}). The spatial adjusting weight map $\boldsymbol{M}_a^j \in \mathbb{R}^{1 \times H \times W}$ can be obtained by a $1 \times 1$ convolution layer after the interaction. Additionally, each input has its own spatial weight map $\boldsymbol{M}_o^j \in \mathbb{R}^{1 \times H \times W}$ extracted from the windows by another $1 \times 1$ convolution layer directly. Thus, the outputs $\boldsymbol{F}^j$ corresponding to the two inputs can be expressed as:
\begin{equation}
    \begin{aligned}
        \boldsymbol{F}^j={\rm Mul}(\boldsymbol{F}_w^j, {\rm Add}(\boldsymbol{M}_o^j, \alpha\boldsymbol{M}_a^j)), j=1,2,
    \end{aligned}
\end{equation}

\noindent
where $\alpha$ is a learnable parameter. Finally, windows are restored as features and final output is obtained by smoothing out possible seams with a layer of $3 \times 3$ convolution.

\noindent
\textbf{Channel Grouping Feature Interaction.}  As spatial feature interaction concentrates on local details, we further introduce channel grouping feature interaction to calibrate global information inspired by \cite{guo2022isdnet}. Firstly, we transfer the input $\boldsymbol{F}_{in}^j$ to channel-wised attention vector $\boldsymbol{A}_o^j \in \mathbb{R}^C$ by applying global average pooling and an MLP layer. Then, the vectors are fed into a feature interaction module, and two adjusting attention vectors $\boldsymbol{A}_a^j \in \mathbb{R}^C$ integrating the two features are obtained through another MLP layer. Similar to the spatial one, the final outputs can be obtained by:
\begin{equation}
    \begin{aligned}
        \boldsymbol{F^j}={\rm Mul}(\boldsymbol{F}_{in}^j, {\rm Add}(\boldsymbol{A}_o^j, \beta\boldsymbol{A}_a^j)), j=1,2
    \end{aligned}
\end{equation}

\noindent
where $\beta$ is a learnable parameter. Since global information is important for both blur~\cite{sahu2019blind} and semantic estimation~\cite{liu2015parsenet}, feature interaction of channel dimension is also essential.

\begin{table*}
  \centering
  \begin{tabular}{@{}lcccccc@{}}
    \toprule
    Method & Group1 & Group2 & Group3 & Group4 & Group5 & Avg.\\
    \midrule
    KernelGAN~\cite{bell2019blind} & 23.10/0.7430 & 23.13/0.7439 & 23.18/0.7449 & 23.16/0.7449 & 23.18/0.7461 & 23.15/0.7446\\
    KOALAnet~\cite{kim2021koalanet} & 27.69/0.8773 & 27.73/0.8768 & 27.60/0.8734 & 27.73/0.8754 & 27.74/0.8760 & 27.70/0.8758\\
    DCLS~\cite{luo2022deep} & 27.89/0.8799 & 27.94/0.8798 & 27.82/0.8760 & 27.91/0.8768 & 27.89/0.8781 & 27.89/0.8781\\
    DAN~\cite{huang2020unfolding} & 27.90/0.8809 & 27.98/0.8808 & 27.83/0.8771 & 27.91/0.8775 & 27.88/0.8791 & 27.90/0.8791\\
    MANet~\cite{liang2021mutual} & \textcolor{blue}{30.16/0.9117} & \textcolor{blue}{30.20/0.9111} & \textcolor{blue}{30.07/0.9095} & \textcolor{blue}{30.07/0.9099} & \textcolor{blue}{30.10/0.9107} & \textcolor{blue}{30.12/0.9106} \\
    CMOS(ours) & \textcolor{red}{32.09/0.9168} & \textcolor{red}{32.08/0.9159} & \textcolor{red}{31.99/0.9145} & \textcolor{red}{31.96/0.9147} & \textcolor{red}{32.01/0.9153} & \textcolor{red}{32.03/0.9154} \\
    \cdashline{1-7}[3pt/3pt]
    Upper Bound & 33.80/0.9309 & 33.78/0.9303 & 33.69/0.9290 & 33.73/0.9301 & 33.74/0.9298 & 33.75/0.9300\\
    \bottomrule
  \end{tabular}
  \caption{Average PSNR/SSIM of different methods for spatially variant blind SR on NYUv2-BSR. Avg. represents the average results on the 5 test groups. The best and second best results are highlighted in \textcolor{red}{red} and \textcolor{blue}{blue} colors, respectively.}
  \label{tab:NYUv2}
\end{table*}
\setlength{\intextsep}{0pt}

\begin{table*}
  \centering
  \begin{tabular}{@{}lcccccc@{}}
    \toprule
    Method & Group1 & Group2 & Group3 & Group4 & Group5 & Avg.\\
    \midrule
    KernelGAN~\cite{bell2019blind} & 28.96/0.8461 & 29.02/0.8475 & 28.88/0.8464 & 28.96/0.8468 & 28.99/0.8477 & 28.96/0.8469\\
    KOALAnet~\cite{kim2021koalanet} & 32.40/0.9173 & 32.45/0.9177 & 32.29/0.9149 & 32.38/0.9166 & 32.40/0.9166 & 32.38/0.9166\\
    DCLS~\cite{luo2022deep} & 32.41/0.9174 & 32.46/0.9176 & 32.28/0.9151 & 32.44/0.9168 & 32.38/0.9166 & 32.39/0.9167\\
    DAN~\cite{huang2020unfolding} & 32.33/0.9162 & 32.38/0.9165 & 32.21/0.9140 & 32.36/0.9156 & 32.30/0.9155 & 32.32/0.9156\\
    MANet~\cite{liang2021mutual} & \textcolor{blue}{34.24/0.9293} & \textcolor{blue}{34.29/0.9294} & \textcolor{blue}{34.16/0.9273} & \textcolor{blue}{34.27/0.9288} & \textcolor{blue}{34.27/0.9285} & \textcolor{blue}{34.25/0.9287}\\
    CMOS(ours) & \textcolor{red}{35.58/0.9388} & \textcolor{red}{35.61/0.9389} & \textcolor{red}{35.50/0.9373} & \textcolor{red}{35.60/0.9385} & \textcolor{red}{35.60/0.9381} & \textcolor{red}{35.58/0.9383} \\
    \bottomrule
  \end{tabular}
  \caption{Average PSNR/SSIM of different methods for spatially variant blind SR on Cityscapes-BSR. Avg. represents the average results on the 5 test groups. Note that, there is no official ground truth semantic maps for the test sets of Cityscapes~\cite{cordts2016cityscapes}, so the upper bound is not available here. The best and second best results are highlighted in \textcolor{red}{red} and \textcolor{blue}{blue} colors, respectively.}
  \label{tab:Cityscapes}
\end{table*}
\vspace{0pt}

\noindent
\textbf{Feature Group Interaction.}  This module is designed to interact spatial or channel features in groups. For spatial interaction, the input size is $C \times H \times W$. We regard the features of each pixel as a group, and the size of the grouped features is 
$N \times D$, where $N = HW$, $D = C$. For channel interaction, the input size is $C$. It will be divided into $N$ groups with length $D$, where $C=ND$. In this way, both spatial and channel inputs can be represented as $\boldsymbol{G}_{i} \in \mathbb{R}^{N \times D}, i=1,2$ after grouping, where $i$ represents two different inputs. Then, we use inner product for feature interaction and get the interactive feature $\boldsymbol{F}_{fuse} \in \mathbb{R}^{N \times N}$,
\begin{equation}
    \begin{aligned}
        \boldsymbol{F}_{fuse}=\boldsymbol{G}_1{\boldsymbol{G}_2}^T.
    \end{aligned}
\end{equation}

\noindent
After that, for spatial interaction, one of the output can be obtained by reshaping $\boldsymbol{F}_{fuse}$ to $H \times W \times N$, and the other can be obtained by reshaping $\boldsymbol{F}_{fuse}$ to $N \times H \times W$. For channel interaction, the two final outputs are the same and can be both obtained by simply flatten $\boldsymbol{F}_{fuse}$.

\subsection{Loss Function}
We use the mean absolute error (MAE) for blur estimation and the cross-entropy (CE) loss for semantic segmentation. As shown in~\cref{fig:main} (a), the auxiliary loss $\mathcal{L}_{1}$ and loss $\mathcal{L}_{3}$ are both MAE,
while the auxiliary loss $\mathcal{L}_{2}$ and loss $\mathcal{L}_{4}$ are both CE, specifically:
\begin{eqnarray}
\mathcal{L}_{1} = \mathcal{L}_{3} = \frac{1}{H \times W} \sum_{i=1}^{H}\sum_{j=1}^{W} \lVert \boldsymbol{B}_{i,j}-\hat{\boldsymbol{B}}_{i,j} \rVert_{1}\\
\mathcal{L}_{2} = \mathcal{L}_{4} = -\frac{1}{H\times W}\sum_{i=1}^{H}\sum_{j=1}^{W}\sum_{c=1}^{C}\boldsymbol{S}_{i,j}^c\log(\hat{\boldsymbol{S}}_{i,j}^c)
\end{eqnarray}

\noindent
where $\hat{\boldsymbol{B}}_{i,j}$ and $\boldsymbol{B}_{i,j}$ denote the estimated blur map and the corresponding ground-truth at position $(i,j)$. Similarly, $\hat{\boldsymbol{S}}_{ij}^c$ and $\boldsymbol{S}_{ij}^c$ represent the estimated semantic map and the ground-truth at position $(i,j)$ of the $c$-th category. $C$ is the number of object categories, and $H, W$ are the height and width of the maps.  We do not adopt a particular loss weighing strategy, but simply sum the losses together,
\begin{eqnarray}
\mathcal{L} = \mathcal{L}_{1} + \mathcal{L}_{2} + \mathcal{L}_{3} + \mathcal{L}_{4}
\end{eqnarray}
\section{Experiments}
\subsection{Experimental Setup}

\noindent
\textbf{Settings of CMOS.}  We select HRNet~\cite{wang2020deep} as our backbone and change the stride of the first two convolutions to 1. This translates to 4 scales of the input LR images (1, 1/4, 1/8, 1/16). The task-specific heads are implemented as two basic residual blocks~\cite{he2016deep}. As for semantic segmentation, we use the official 40 classes for NYUv2-BSR and 19 classes for Cityscapes-BSR. All our experiments are conducted with the pre-trained ImageNet weights.

\noindent
\textbf{Settings of Non-Blind SR.}  For non-blind SR, we use RRDB-SFT proposed in \cite{liang2021mutual}. To feed both blur and semantics into it, we use a GIA module. Finally, we fine-tune RRDB-SFT on blur and semantic maps estimated by CMOS. The loss between SR and HR images is also MAE.

\noindent
\textbf{Implementation Details.} The image sizes are selected as $640 \times 480$ for both NYUv2-BSR and Cityscapes-BSR. We augment the training data by scaling with a randomly selected ratio in $\left\{1, 1.2, 1.5\right\}$ and the blur values are divided by the ratio. We also flip the training samples with a possibility of 0.5. Adam optimizer~\cite{kingma2014adam} with $\beta_1=0.9$ and $\beta_2=0.99$ is used to train the model for 700 epochs, with a batch size of 8. The learning rate is initialized as 0.0001 and a cosine learning rate schedule with 10 warm-up epochs is adopted. Implemented with PyTorch, it takes about 28 hours to train CMOS on an RTX 3090 GPU.

\noindent
\textbf{Evaluation Metrics.}  For blur estimation, we use PSNR and SSIM~\cite{wang2004image}. For semantic segmentation, we use mIoU. For the final SR images generated by RRDB-SFT with the blur and semantic maps estimated by CMOS, we compare PSNR/SSIM on the Y channel of YCbCr space.

\begin{table}
  \centering
  \renewcommand{\arraystretch}{1.0}
  \begin{tabular}{@{}lcc@{}}
    \toprule
    Datasets & PSNR $\uparrow$ & SSIM $\uparrow$ \\
    \midrule
    IVA & 19.50 & 0.6840 \\
    NYUv2-BSR & \textbf{30.12} & \textbf{0.9106} \\
    \bottomrule
  \end{tabular}
  \caption{Importance of using space-variant blur for training. IVA stands for the NYUv2 dataset with only space-invariant blur.}
  \label{tab:dataset importance}
\end{table}

\subsection{Comparison with the State-of-the-Arts}
We compare CMOS with existing blind SR models: KernelGAN~\cite{bell2019blind}, KOALAnet~\cite{kim2021koalanet}, DCLS~\cite{luo2022deep}, DAN~\cite{huang2020unfolding}, MANet~\cite{liang2021mutual} and the upper bound model (RRDB-SFT given ground-truth blur and semantic maps). We retrained all the comparison methods on NYUv2-BSR and Cityscapes-BSR using their official implementations and settings. KernelGAN is an unsupervised method which trained solely on the LR image at test time. DCLS and DAN are end-to-end methods for space-invariant blur, while KOALAnet and MANet are two-stage methods for space-variant blur. Since we use the non-blind SR model proposed in MANet (\ie RRDB-SFT), we apply same settings to ensure the fairness.

\begin{figure*}[htp]
    \centering
    \includegraphics[width=16cm]{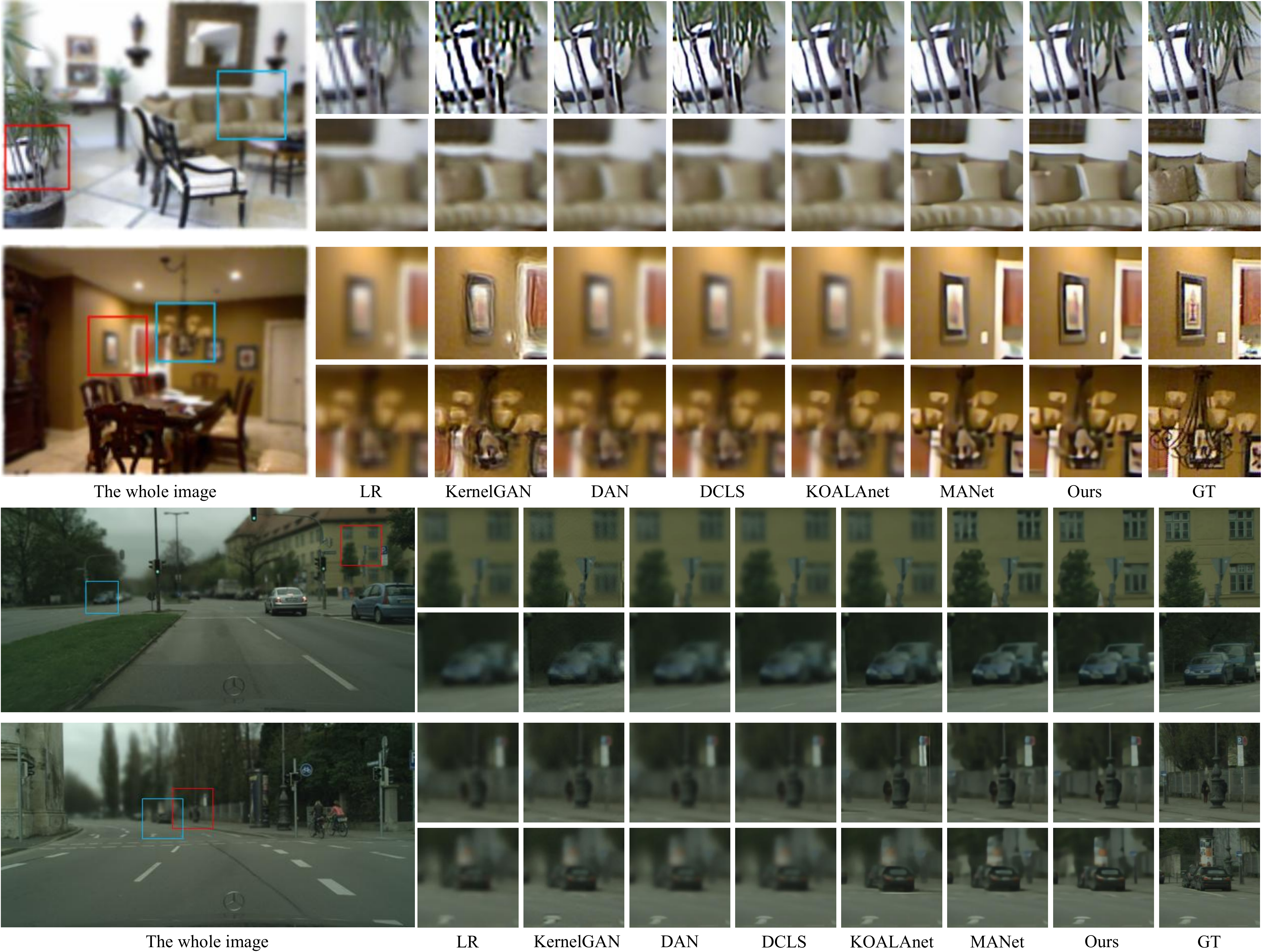}
    \caption{Qualitative comparisons between different SR methods on spatially variant blur (out-of-focus). The first two pictures are from NYUv2-BSR and the last two are from Cityscapes-BSR. (Please zoom in for better view.)}
    \label{fig:nyucity}
\end{figure*}
\vspace{0pt}

\noindent
\textbf{Quantitative comparison.}  As shown in Tab.~\ref{tab:NYUv2} and Tab.~\ref{tab:Cityscapes}, CMOS leads to the best performance for different test groups in both the two proposed datasets. Notably, methods that estimate only one blur kernel for an image (\ie, KernelGAN, DCLS, and DAN) all suffer from severe performance drop when the real kernels are spatially variant. Although KOALAnet estimates different kernels for different image pixels, it does not include any special handling for space-variant properties and also produces unfavorable results. MANet takes the locality of blur into account, so it performs relatively better. By contrast, the proposed model CMOS effectively utilizes semantic information to help with spatially variant blur estimation and non-blind SR, outperforming MANet by large margins.

\noindent
\textbf{Qualitative comparison.}  We present several representative visual samples in Fig.~\ref{fig:nyucity}. It can be observed that our CMOS outperforms previous approaches in both removing blur and avoiding artifacts. Other methods may either produce ringing artifacts (especially KernelGAN), or fail to restore texture details, leading the patches still blurry.

\subsection{Ablation Study}

All the experiments in this section use NYUv2-BSR for training, and the metrics (\ie PSNR, SSIM and mIoU) refer to the mean value across the 5 test sets (Sec.~\ref{sec:sec3}).

\noindent
\textbf{Importance of Using Space-Variant Blur for Training.} According to \cite{liang2021mutual}, because of the moderate receptive field, MANet can handle spatially variant cases even if it is trained on spatially invariant blurred images. But we believe that it is necessary to use the images containing space-variant blur for training. To prove it, we trained two MANet models: one on the proposed NYUv2-BSR dataset, and the other on the space-invariant blurred images generated from the NYUv2 dataset. The comparison results are shown in the Tab.~\ref{tab:dataset importance}. Apparently, training on spatially variant blurred images can increase PSNR and SSIM of SR images dramatically by 10.62 dB $\uparrow$ and 0.2266 $\uparrow$, respectively. This indicates that maintaining consistency in image blur types during the training and testing phases is crucial.

\begin{figure*}[htp]
    \centering
    \includegraphics[width=16cm]{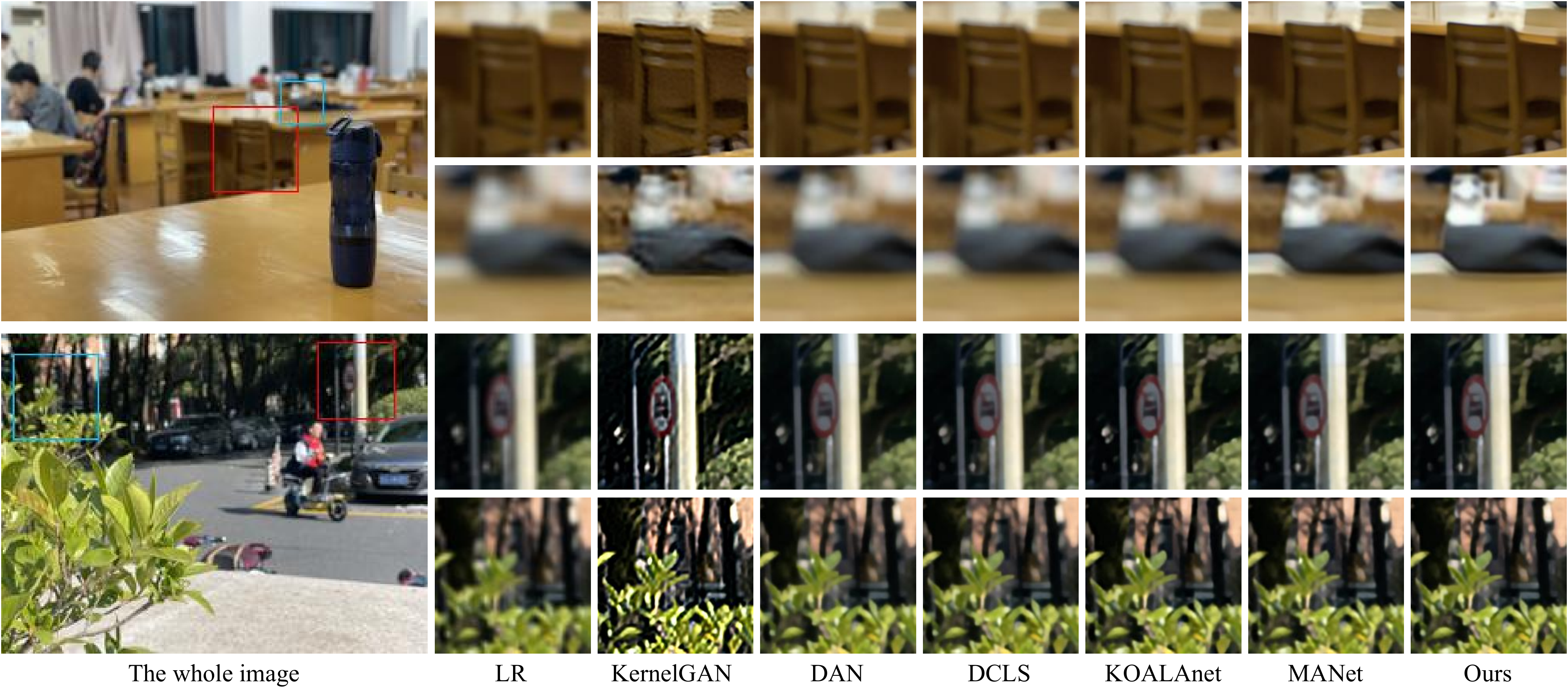}
    \caption{Visual results on real-world images for scale factor 4. The first picture of the indoor scene uses the model trained on NYUv2-BSR, and the second picture of the outdoor scene uses the model trained on Cityscapes-BSR. (Please zoom in for better view.)}
    \label{fig:in the wild}
\end{figure*}

\begin{table}
  \centering
  \renewcommand{\arraystretch}{1.0}
  \begin{tabular}{@{}lccc@{}}
    \toprule
    Method & PSNR $\uparrow$ & SSIM $\uparrow$ & mIoU $\uparrow$\\
    \midrule
    Ours w/o GIA & 23.21 & 0.8312 & 32.15\\
    Ours w/ F  & 23.42 & 0.8314 & 33.04\\
    Ours w/ F+C & \underline{24.24} & \underline{0.8336} & \textbf{36.25} \\
    Ours w/ F+C+S (GIA) & \textbf{24.52} & \textbf{0.8340} & \underline{35.61}\\
    \bottomrule
  \end{tabular}
  \caption{Effectiveness of GIA. Note that these are the intermediate results, and PSNR/SSIM refer to the blur maps rather than the final SR mages. mIoU evaluates the effect of semantic estimation.}
  \label{tab:GIA}
\end{table}

\noindent
\textbf{Effectiveness of GIA Module.}  We take out the components, \ie, flow-based upsampling (F), channel interaction (C) and spatial interaction (S), of GIA to verify validity. We record the best PSNR and mIoU models individually. As shown in Tab.~\ref{tab:GIA}, using only flow-based upsampling improves the results slightly, and when combined with channel interaction, the performance can be significantly enhanced. Furthermore, utilizing all three components, \ie, the complete GIA module, can yield even greater improvements.

\noindent
\textbf{Effectiveness of Semantic Information in SR.}  In order to illustrate that the semantic information is conducive to SR, we ablate it and only input blur maps into RRDB-SFT. It is worth noting that we use the ground truth blur and semantic maps here. As shown in Tab.~\ref{tab:semseg importance}, adding semantic maps improves the PSNR (+0.34 dB $\uparrow$) and SSIM (+0.0022 $\uparrow$) of the final SR results. We hold the opinion that semantic information may allow the network to take advantage of textural features of related objects it has learned about, and sharp semantic edges may also be helpful in SR. 

\begin{table}
  \centering
  \renewcommand{\arraystretch}{1.0}
  \begin{tabular}{@{}lcc@{}}
    \toprule
    Method & PSNR $\uparrow$ & SSIM $\uparrow$ \\
    \midrule
    RRDB-SFT w/o semseg & 33.41 & 0.9278 \\
    RRDB-SFT w/ semseg & \textbf{33.75} & \textbf{0.9300} \\
    \bottomrule
  \end{tabular}
  \caption{Importance of using semantic information in SR. The ground-truth blur and semantic maps are used in this experiment.}
  \label{tab:semseg importance}
\end{table}
\setlength{\intextsep}{0pt}

\begin{table}[ht]
\centering
\renewcommand{\arraystretch}{1.2}
\begin{tabular}{@{}lccr@{}}
\toprule
\multirow{2}{*}{Method} & \multicolumn{2}{c}{Intermediate Results} & \multicolumn{1}{c}{SR Results}\\
\cmidrule(lr){2-3}\cmidrule(lr){4-4}
& PSNR/SSIM $\uparrow$ & mIoU $\uparrow$ & PSNR/SSIM $\uparrow$ \\
\midrule
Single Task & \textbf{24.58}/\textbf{0.8393} & 33.95 & 30.75/0.9134\\
CMOS (Ours) & 24.52/0.8340 & \textbf{35.61} & \textbf{32.03}/\textbf{0.9154}\\
\bottomrule
\end{tabular}
\caption{Effectiveness of MTL. PSNR/SSIM and mIoU of the intermediate results refer to the blur maps and the semantic maps.}
\label{tab:mtl}
\end{table}

\noindent
\textbf{Effectiveness of Multi-task Learning (MTL).} To demonstrate the effectiveness of MTL, firstly, we make separate predictions for blur and semantic maps and compared them with the CMOS results. Secondly, We compare the SR results achieved by solely utilizing the estimated blur maps versus employing both the estimated blur and semantic maps. As shown in~\cref{tab:mtl}, joint estimation improves the results of semantic segmentation (mIoU +1.66$\uparrow$), albeit with a slight decrease in the performance of blur estimation. But in general, MTL can improve the PSNR/SSIM of the final SR results by +1.28$\uparrow$/+0.002$\uparrow$, which proves that semantics is useful to the overall SR process.

\noindent
\textbf{Importance of the Auxiliary Supervision.}  We ablate the auxiliary supervision in CMOS to see if it is necessary for our framework. As shown in Tab.~\ref{tab:supervision}, without the auxiliary supervision in the multi-scale structure, although there is a slightly increase in SSIM, PSNR and mIoU droped by 0.38 $\downarrow$ and 0.24\% $\downarrow$, respectively. Therefore, auxiliary supervision can improve the performance of CMOS on the whole.

\subsection{Experiments on Real-Wrold SR}

As there is no ground-truth for real images, we only compare visual results of different methods. As shown in Fig.~\ref{fig:in the wild}, similar to the results on our datasets, KernelGAN still generate ringing artifacts, especially in the ourdoor scene. DAN, DCLS and KOALAnet all produce blurry results, while MANet performs slightly better. In comparision, CMOS can produce realistic and natural textures, and the results are the clearest.

\begin{table}
  \centering
  \renewcommand{\arraystretch}{1.0}
  \begin{tabular}{@{}lccc@{}}
    \toprule
    Methods & PSNR $\uparrow$ & SSIM $\uparrow$ & mIoU $\uparrow$ \\
    \midrule
    CMOS w/o AS & 24.14 & \textbf{0.8347} & 35.37\\
    CMOS w/ AS & \textbf{24.52} & 0.8340 & \textbf{35.61}\\
    \bottomrule
  \end{tabular}
  \caption{Importance of the auxiliary supervision (AS) in CMOS.}
  \label{tab:supervision}
\end{table}

\section{Conclusion}
In this paper, we introduce out-of-focus blur to SR and propose two new datasets: NYUv2-BSR and Cityscapes-BSR. Besides, we further propose a novel model CMOS to estimate the blur and semantic maps simultaneously. By incorporating semantics, we can restore finer SR results. GIA modules is used to achieve effective feature interaction in both spatial and channel dimensions. Extensive experiments on proposed datasets and real-world images demonstrate that our model can achieve SOTA performance in blind SR when integrated with existing non-blind models.

\noindent
\textbf{Acknowledgments:} This work is supported by the National Natural Science Foundation of China (61836015).

{\small
\bibliographystyle{ieee_fullname}
\bibliography{PaperForArxiv}
}

\end{document}